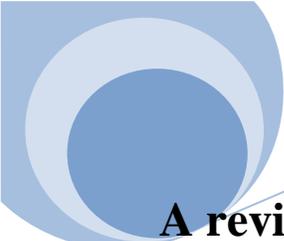

# A review on handwritten character and numeral recognition for Roman, Arabic, Chinese and Indian scripts.

Aini Najwa Azmi
Faculty of Computing,
Universiti Teknologi Malaysia,
81310 UTM Skudai,
Johor, Malaysia.
aininajwa.azmi@gmail.com

Dewi Nasien
Faculty of Computing,
Universiti Teknologi Malaysia,
81310 UTM Skudai,
Johor, Malaysia.
dewinasien@utm.my

Siti Mariyam Shamsuddin
Faculty of Computing,
Universiti Teknologi Malaysia,
81310 UTM Skudai,
Johor, Malaysia.
mariyam@utm.my

*Abstract*— There are a lot of intensive researches on handwritten character recognition (HCR) for almost past four decades. The research has been done on some of popular scripts such as Roman, Arabic, Chinese and Indian. In this paper we present a review on HCR work on the four popular scripts. We have summarized most of the published paper from 2005 to recent and also analyzed the various methods in creating a robust HCR system. We also added some future direction of research on HCR.

## I. INTRODUCTION

Handwritten Character Recognition (HCR) is an automation process and can improve the interface between man and machine in a lot of applications. Generally, handwritten character recognition is classified into two types which are offline and online handwritten character recognition methods. In the offline recognition, the writing is usually captured optically by a scanner and the completed writing is available as an image. But, in the on-line system the two dimensional coordinates of successive points are represented as a function of time and the order of strokes made by the writer are also available [1, 2]. HCR system is a very complex and challenging problems because of variability on size, writing style of hand-printed characters, and duplicate pixels caused by a hesitation in writing or interpolate non-adjacent consecutive pixels caused by fast writing [3].

Some practical applications of HCR systems are: processing cheques without human involvement, reading aid for the blind, automatic text entry into the computer for desktop publication, library cataloguing, health care, and ledgering, automatic reading of city names and addresses for postal mail, document data compression, natural language and processing investigation forms or the automatic reading of postal addresses [4,5].

Generally, in HCR system consists three stages which are pre-processing, feature extraction and classification. The first step of processing usually consists of image enhancement and converting the grey level image to binary image as required in image pre-processing. After converting image from the gray-scale image to binary format, the thresholding technique is used to separate the useful front pixels from the background pixels. Noise reduction is performed before or after binarization, which identifies and corrects the noise pixels. These sorts of techniques are based on image filtering theory and mathematical morphology. Furthermore, a normalization step normalizes the handwriting sample images from varied stroke width. The methods generally apply to binary images and normalize the strokes width to single pixel thickness. Noise is a term that normally used for non information-bearing variability that is introduced by one or more physical processes, such as scanning, faxing, writing style, presence or absence of ruled lines, crumpling and folding. This is usually happened in off-line HCR system. The goal of pre-processing is to minimize noise before the image is further processed to next stage which is extracting the features. Feature extraction plays an important role in handwriting recognition. In HCR process, a text image must be either processed by feature extraction after image pre-processing. The selected features will be the inputs for classifier and perform matching. Features are the information passed to the classifier such as pixels, shape data or mathematical properties. Classifier is used to rate the efficiency of the system.

This paper is divided to four sections. Section I describes introduction. Section II describes related work on HCR system Roman, Arabic, Chinese and Indian scripts. Section III describes related work on HCR system includes pre-processing, feature extraction and classification. Section IV shows conclusion of the whole content.

## II. RELATED WORK ON HANDWRITTEN CHARACTER RECOGNITION

In this section, we report various HCR systems for Roman, Arabic, Chinese and Indian scripts. This section is divided into four parts:
A. Roman Handwritten Character and Numeral Recognition,
B. Arabic Handwritten Character and Numeral Recognition,
C. Chinese Handwritten Character and Numeral Recognition,
D. Indian Handwritten Character and Numeral Recognition.

### A. *Roman Handwritten Character and Numeral Recognition*

Table 1 below discussed the stage of HCR includes pre-processing, feature extraction and classification for Roman Handwritten and Numeral script.

Table 1 HCR system for Handwritten Roman

| Authors | Pre-processing | Features extraction/ Classification | Details of result/Description |
|---|---|---|---|
| Pradeep *et al.* (2011)[1] | Segmentation, binarization, noise removal | Diagonal Feature Extraction Method/ Neural Network | Result showed high on diagonal compared to vertical and horizontal feature extraction. The rate was 98.54% |
| Zhang *et al.* (2005) [6] | Binarization and skeletonization | Complex Wavelet Features/ Artificial Neural Network (ANN) | Highest result was 99.12% for the mentioned feature set using single classifier. But combination of 3 classifier showed better result which was 99.25% |
| Wang & Sajjahar (2011) [2] | Binarization, noise reduction, segments lines, and normalization | Polar transformed images, Zone based feature extraction/ Support Vector Machines (SVM) | Highest result was 86.63% by using polar coordination for the Kernel Function |
| Verma *et al.* (2004) [3] | Dehooking | Structural features-the change of writing direction, and zoning information to create a single global feature vector/ Neural Network | Highest test result was 86.63% for digit by using 40 hidden units. |
| Choudhary *et. al* (2012) [7] | Noise removal and resizing | Vertical, Horizontal, Left Diagonal and Right Diagonal directions/ Neural Network | The average recognition accuracy was 95.33%. |
| Rani & Meena (2011) [8] | Binarization, resizing, thinning | Cross-corner feature extraction method/ Back propagation neural network (BPN) | Cross-corner provided substantial increase in accuracy without increasing much in feature space size. Result on dataset 1 was 97% while dataset 2 is 93% |
| Vamvakas *et al.* (2010) [9] | Normalization | Zoning based features, upper and lower character profile projections features, left and right character profile projections features, distance based features/ Neural Network and SVM | MNIST Digit Database showed highest test result which was 98.08% for single stage classification. Two-stage classification recognition rate showed 99.03% |
| Li *et al.* (2012) [10] | Normalization | direction string and nearest neighbor matching/ Nearest neighbor matching based classifier | Recognition accuracy reached almost 99% averagely. |
| Yang *et al.* (2011) [11] | noise reduction | structural features and the statistical features/ Back propagation neural network (BPN) | In this method, structure and statistical features were extracted and combined to form the testing features and yielded 100% accuracy. |
| Chel *et al.* (2011) [12] | Binarization and segmentation | Transition Feature, Sliding Window Amplitude Feature, Contour Feature/ Neural Network | Maximum result was 92.32%. Though the degree of recognition was good but it may further be improved by lexicon matching technique |
| Gomathi Rohini *et al.* (2013) [13] | Binarization and segmentation | Transition features/ Neural Network | Classification achieved 92%. It showed that using simple transition feature, the segmentation rate was better. |
| Fink and Flotz (2005) [14] | Not specify | Principle Component Analysis (PCA) based features, | Lowest Character Error Rate (CER) was 26% |

| Authors | Pre-processing | Features extraction/Classification | Details of result/Description |
|---|---|---|---|
| | | Discrete Wavelet Transform (DWT) features, and geometrical features/ Hidden Markov Model (HMM) | |
| Uchida & Liwicki (2010) [15] | Not specify | Speeded Up Robust Features (SURF)-Upgraded SIFT/ Neural Network | They achieved 93.8% by using a small part (about 1/20 of the character size) as the unit area of local feature description. |

There are some papers that discussed about Freeman Chain Code (FCC) in extracting the character features. The table below are discussing some papers that used FCC in their system. There are two directions of chain code, namely 8-neighborhood and 4-neighborhood.

Table 2 FCC in HCR

| Authors | Pre-processing | Features extraction/Classification | Details of result/Description |
|---|---|---|---|
| Bayoudh et al. (2007) [16] | Not specify | Chain code features (16-FCC)/ Radial Basis Function Neural Network and SVM | Writer-dependent recognition rates and their standard deviation depending on the number of used original characters compared to reference rates using 10 or 30 characters per class for RBFN and SVM classifiers. |
| Lee et al. (2010) [17] | Not specify | Chain code features (Cyclic FCC Histogram, 8FCC)/ Sigmoid RBF + Growing/Pruning | Without slant and skew deformation, CCH feature produced a recognition rate of 90.1%, and cyclic-CCH was 91.4%. With slant and skew deformation, the recognition rate of CCH feature decreased to 60.3%, and cyclic-CCH decreased to 67.9%. |
| Hasan et al. (2009) [18] | Thinning | Chain code features (8FCC)/ Neural Network | Proposed PSO performs better than the proposed DE. This can be seen by comparing their average, max, and standard deviation result. The computation time decreased twice. |

## B. Arabic Handwritten Character and Numeral Recognition

Table 2 below discussed the stage of HCR includes pre-processing, feature extraction and classification for Arabic Handwritten and Numeral script.

Table 3 HCR system for Arabic Handwritten and Numeral script

| Authors | Pre-processing | Features extraction/Classification | Details of result/Description |
|---|---|---|---|
| Nemouchi et al. (2012) [19] | Thresholding, smoothing, skeletonizing and contouring | Structural(like strokes, concavities, end points, intersections of line segments, loops, stroke relations) and statistic (zoning, invariants moments, Fourier descriptors, Freeman chain code) features/ Fuzzy C-Means algorithm (FCM), the K-Means algorithm, the K Nearest Neighbor algorithm (KNN) and a Probabilistic Neural Network (PNN) | Combination of 3 features yielded 70% accuracy |
| Ahmed et al. (2012) [20] | Smoothing and skew correction | HMM based feature, Zoning of pixel and statistical features (zoning of the character array (i.e., dividing it into over-lapping or non-overlapping regions, computing the moments of the black pixels of the character, the n-tuples of black or white or joint occurrence, the characteristic loci, and crossing distances)/ Hidden Markov Model (HMM) | Using this new feature extraction algorithm, they obtained 98% of accuracy, a significant improvement compared to the best result of 81.45 % using the hierarchical features. |
| Al-Khateeb et al. (2011) [21] | Removing noise, image enhancement, and segmentation | Structural which is geometrical and topological features (strokes, endpoints, loops, dots and their position related to the baseline) and statistical features(statistical distribution of pixels and describing the characteristic measurements of a pattern, which in-clude zoning, | The best result was generated by using fusion of multiple HMMs which was 95.15% |

| Authors | Pre-processing | Features extraction/ Classification | Details of result/Description |
|---|---|---|---|
| | | density distribution of pixels that counts the ones and zeros, moments )/ Hidden Markov Model (HMM) | |
| Pechwitz et al. (2012) [22] | noise reduction, segmentation, and binarization | Calculate aspect ratio from skeleton graph/ Hidden Markov Model (HMM) | The best recognition rate was about 92 % with segmentation. |
| Likforman-Sulem et al. (2012) [23] | Segmentation | Structural and statistic features/ Neural Network | Combination context-independent+ grapheme MLP-HMM showed highest recognition rate which was 89.42% |
| Lawal et al. (2010) [24] | Normalization and segmentation | Chain code features (8-FCC)/ Neural Network | Average recognition rate of 99.03% was obtained |
| Kessentini et Al. (2012) [25] | Normalization, contour smoothing, and baseline detection | Directional density and (black) pixel densities features/ Hidden Markov Model (HMM) | A two-level decoding algorithm was proposed to reduce the complexity of the decoding step and significantly speed up the recognition process while maintaining the recognition accuracy |
| El Abed & Margner (2007) [26] | Binarization, word segmentation, and noise reduction | 1) Sliding Window with Pixel Feature 2) Skeleton Direction-based Features 3) Sliding Window with Local Features/ Hidden Markov Model (HMM) | We achieved recognition rates of up to 89% on word level using the skeleton based method for baseline estimation and skeleton direction features. |
| Moradi et al. (2009) [27] | Binarization | 1) A statistical approach is used for representing the spatial distribution of the pixel values of binary image 2) Count the number of intersections along middle vertical ray and divide the pictures to eight sections (4 vertical,4 horizontal) 3) Elastic Meshing Directional Feature Extraction/ Multi Layer Perceptron (MLP) | Recognition system rate for testing data was 97.62% by using 16 hidden layer of classifier. |
| Alaei et al. (2010) [28] | Normalization | Undersampled bitmaps and directional chain code information/ Support Vector Machines (SVM) | Obtained the best recognition rate of 96.17% when 196 directional features with overlapping window-map are used. |

## C. Chinese Handwritten Character and Numeral Recognition

Table 3 below discussed the stage of HCR includes pre-processing, feature extraction and classification for Arabic Handwritten and Numeral script.

Table 4 HCR system for Handwritten Chinese

| Authors | Pre-processing | Features extraction/ Classification | Details of result/Description |
|---|---|---|---|
| Wang et al. (2012) [29] | Segmentation | Local stroke direction histogram feature/ Modified Quadratic Discriminant Function (MQDF), Nearest Prototype Classifier (NPC) | Recognition accuracies with ground-truth line segmentation using MQDF classifier yielded highest result 94.28%. Error can be reduced by optimizing the techniques in all steps which will be future work for this paper. |
| Liu et al. (2013) [30] | Normalization | Gradient direction features/ Modified Quadratic Discriminant Function (MQDF) | Result showed 12-direction feature had a better tradeoff between accuracy and complexity compare to 8-directional and 16-direction. Lowest error rate was 1.73% |
| Su et al. (2009) [31] | Thinning | Orientation difference, width comparison, curvature variation, domain knowledge/ Bayesian | High result achieved 98.1% for ambiguous zone detection and 95.63% for stroke extraction. |
| Ma & Leedham (2007) [32] | noise reduction, time normalization and size normalization | Aspect ratio calculation/ Neural network | Experiment showed that proposed method recognized vocalized outline and Renqun shortform which were written following the new writing rule |

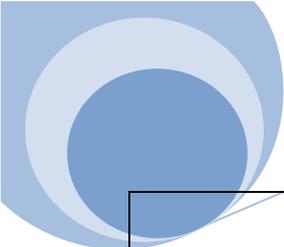

| Authors | Pre-processing | Features extraction/ Classification | Details of result/Description |
|---|---|---|---|
| | | | with classification rate of 83% and 84.07%. Future work will concentrate on detecting hooks and post processing. |
| Ni *et al.* (2012) [33] | Not specify | Haar-like features (Upright and tilted features)/ Cascade Classifier | Proposed method achieved detection rates of 94.29% (method 1) and 96.14% (method 2) |
| Cheng Lin (2006) [34] | Binarization and normalization | Discriminative feature extraction (DFE) and discriminative learning quadratic discriminant function (DLQDF)/ Modified Quadratic Discriminant Function (MQDF) | Compared to the modified quadratic discriminant function (MQDF) with Fisher discriminant analysis, the error rates on two test sets were reduced by factors of 29.9% and 20.7%, respectively. |
| Zhiyi *et al.* (2009) [35] | Normalization | SIFT feature, Gabor feature and gradient feature (Sobel Operators)/ Modified Quadratic Discriminant Function (MQDF) | Experiments using MQDF classifier show our feature's effectiveness with a recognition rate of 97.868%, which outperforms original SIFT feature and two traditional features, Gabor feature and gradient feature. |
| Lee *et al.* (2009) [36] | Cropping and normalization | X-Y graphs decomposition and Haar wavelet/ Neural Network | Experimental results have proved the efficiency of our proposed method and it is superior to other representative traditional feature extraction schemes with high recognition rate of 95.5%, despite of small dimensionality between 64 (inclusive) and 128 (exclusive) and less processing time. |

### D. Indian Handwritten Character and Numeral Recognition

Table 5 below discussed the stage of HCR includes pre-processing, feature extraction and classification for Arabic Handwritten and Numeral script.

Table 5 HCR system for Handwritten Indian

| Authors | Pre-processing | Features extraction/ Classification | Details of result/Description |
|---|---|---|---|
| Bhattacharya *et al.* (2012) [37] | Binarization, Size normalization, Noise cleaning, Headline truncation | Chain code computation, gradient feature and pixel count feature generation/ Modified Quadratic Discriminant Function (MQDF) | Overall recognition accuracy of the proposed scheme is 95.84 %. Future works will improve the pre-processing stage. |
| Desai (2010) [38] | Normalization, smoothing, skew correction | Vector Distance based feature (horizontal, vertical and 2 diagonal)/ Neural Network | This work has achieved approximately 82% of success rate for Gujarati handwritten digit identification. |
| Bhattacharya *et al.* (2006) [39] | Smoothing, Binarization, and Removal of Extra Long Headline | Chain code histogram features/ Multilayer Perceptrons (MLPs) | Final recognition accuracies on the training and the test sets are respectively 94.65% and 92.14%. |
| Shanti & Duraiswamy (2009) [40] | Thinning | Pixel density calculation/ SVM | The system has achieved a very good recognition accuracy of 82.04% on the handwritten Tamil character database. The recognition accuracy of the individual characters can be further improved by combining the multiple classifiers. |
| Toselli *et al.* (2007) [41] | repeated points elimination, noise reduction, writing speed normalization and size normalization. | Time-domain features, frequency domain/ Neural Network | The final classification result of proposed system was around to 9% of error rate. |

| Reference | Pre-processing | Feature extraction/ Classification | Results |
|---|---|---|---|
| Rajashekara-radhya & Ranjan (2009) [42] | normalization and thinning | zone based hybrid approach/ Recognition Nearest Neighbor Classifier (NNC) | Obtained 97.55%, 94%, 92.5% and 95.2% recognition rate for Kannada, Telugu, Tamil and Malayalam numerals respectively. |
| Chacko et al. (2011) [43] | Noise reduction | Wavelet features, chain code features/ ANN | Classifier combination gave good recognition accuracy at level 6 of the wavelet decomposition. |
| Pal et al. (2008) [44] | Binarization | Directional features/ Quadratic classifier based scheme | A five-fold cross validation technique was used for result computation, and we obtained 90.34%, 90.90%, and 96.73% accuracy rates from Kannada, Telugu, and Tamil characters, respectively, from 400 dimensional features |
| Reddy et al. (2008) [45] | Segmentation | 1) polynomial coefficients of the poly- nomial fitted to the plot of distance and angle 2) coefficients of the spline curve fitted onto the points determined by the segmentation algorithms/ SVM | The feature vector was fed to the SVM classifier and it indicated an efficiency of 68% using the polynomial re-gression technique and 74% using the spline fitting method |
| Rajput & Horakeri (2011) [46] | Noise cleaning, binarization | Boundary-based descriptors, namely, crack codes and Fourier descriptors/ K-NN and SVM | The mean performance of the system with these two shape based features together were 91.24% and 93.73% for K-NN and SVM classifiers, respectively, demonstrating the fact that SVM performs better over K-NN classifier |
| Reddy (2012) [47] | Normalization, smoothing, linear interpolation and re-sampling | Vertical and horizontal projection profiles (VPP-HPP), zonal discrete cosine transform (DCT), chain-code histograms (CCH) and pixel level values/ HMM | The combined online and offline system exhibits improved performance over the individual approaches yielded 99.3%. |
| Sharma & Jhajj (2011) [48] | Normalization | Zoning, Directional Distance Distribution (DDD) and Gabor methods/ SVM | Gabor with SVM (Polynomial kernel) gives the best results of all the combinations of feature extraction methods and classification methods yielded 74.29%. Reasons of failure were low quality of images, distorted images and almost similar between characters. |
| Arora et al. (2008) [49] | Thinning, generating one pixel wide skeleton of character image and segmenting the image into 16 segments | Intersection, shadow feature, chain code histogram and straight line fitting features/ Multi Layer Perceptron (MLP) | Obtained 92.80% accuracy for off-line handwritten Devnagari character recognition system |

### E. SUMMARY OF APPROACHES

The most common approaches in recognizer system are Hidden Markov Models (HMMs), Artificial Neural Networks (ANNs) and Support Vector Machine (SVM). But, some of combined or multiple classifier showed excellent result. Variety of techniques in feature extraction has been discussed and it shows that type of scripts also influence the accuracy. Example in one of Indian scripts which is Gurmukhi [49], was having problem in the physical of the character itself that made their accuracy low. Arabic and Chinese scripts also have their own difficulties in extracting features. In Arabic, the dots on, below and in between of the characters gave a lot of challenges to the researchers to solve it while Chinese characters have a lot of strokes that differs between one writer to another. Last but not least, Roman characters also have their own physicality. Slant handwriting is very difficult to recognize that needs proper pre-processing stage to correct it before extracting the features. Continuous writing needs segmentation to isolate the characters.

## F. CONCLUSION

In this paper, we have reported various works on HCR systems in four popular scripts which are Roman, Arabic, Chinese and Indian. We have organized the review according to the type of the scripts. We have reported the research trends, discussed the techniques being used in the modern HCR systems, and the difficulties occurred in the researches. Besides, the accuracy and future work also discussed in this paper.